\documentclass[cameraready]{Interspeech}

\title{Positional Encoding in the Context of \\ Memristor-Based Analog Computation for Automatic Speech Recognition}

\author[affiliation={}, orcid=0009-0002-4721-3479, correspondingauthor]{Benedikt}{Hilmes}
\author[affiliation={}, orcid=0009-0007-0704-6625]{Nick}{Rossenbach}
\author[affiliation={}, orcid=0000-0003-2839-9247]{Ralf}{Schlüter}

\address{
   Machine Learning and Human Language Technology Group, Faculty of Computer Science,\\ RWTH Aachen University, Aachen, Germany \\
    Apptek GmbH, Aachen, Germany
}

\email{\{hilmes, rossenbach, schlueter\}@ml.rwth-aachen.de \vspace{-10cm}}


\usepackage{comment}
\usepackage[normalem]{ulem}

\usepackage{makecell}
\usepackage{multirow}
\usepackage{cleveref}

\usepackage{glossaries}
\newacronym{asr}{ASR}{automatic speech recognition}
\newacronym{vmm}{VMM}{vector-matrix-multiplication}
\newacronym{ctc}{CTC}{Connectionist-Temporal-Classification}
\newacronym{dac}{DAC}{digital-to-analog conversion}
\newacronym{adc}{ADC}{analog-to-digital conversion}
\newacronym{pe}{PE}{positional encoding}

\begin{document}

\maketitle


\begin{abstract}
Memristors provide a new chance for resource-efficient computation of neural models for natural language processing by enabling analog execution of vector-matrix-multiplication.
Yet, computations on these devices are currently subject to larger distortion, both in weight programming and execution. 
In this work, we identify large output values of transformed positional encodings to cause major degradation within analog-to-digital conversion (ADC) as part of memristor-based computation.
By adjusting the proportion of weight and precision bits of the ADC of specific memristor layers, we reduce the degradation of the execution by $\sim50$\% relative, while keeping the estimated energy consumption stable. 
Additionally, we investigate scenarios where the ADC cannot be modified. 
In that case the degradation can be reduced by $\sim30$\% relative after removing encoding-related linear transformations.
\end{abstract}

\section{Introduction}
The increase in model size of modern day natural language processing models pushes the need for efficient computing.
One family of device types that offer efficient computation of \gls{vmm} based operations is so-called memristors. 
With physically programmable resistance states within a crossbar array, such devices can compute \gls{vmm} operations in the analog domain.
Yet, the devices are still in an early stage of development and exhibit significant stochastic behavior, both during programming and execution \cite{Wouters2016}.
Combining a high number of memristors for larger matrices might cause the inaccuracies to ramp up and affect the performance.
This means that especially for large neural networks, where a huge number of memristive devices might be required, the model behavior needs to be studied, to avoid large performance drops.
There are a number of works that investigate the application of memristors both through simulation and small-scale real-life experiments, e.g.\ on datasets such as MNIST \cite{Wan2022, Huang2024, He2026, 9108292, doi:10.1126/science.adf5538, souto2024neuromorphiccircuitsimulationmemristors}.
A recent publication offers a first study related to applying the technology for speech processing by simulating the execution of Conformer-based \gls{asr} models \cite{rossenbach25_interspeech} on memristor hardware.
Yet, some aspects relevant for state-of-the-art speech recognition were missing.
In particular, relative \glspl{pe} in the self-attention, which play an important role in speech modeling, were not part of the model in their analysis.
While it is known that \glspl{pe} are crucial for good performance, we found that they are particularly relevant when operating in low-precision environments.

With the introduction of the Transformer architecture \cite{vaswani2017attention} \glspl{pe} became relevant, because the self-attention mechanism lacks any positional information.
In the initial work absolute \glspl{pe} were added to the input to give positional context. 
An alternative was proposed in \cite{shaw2018self}, where the fixed absolute \glspl{pe} were replaced by relative ones that can be learned during training.
This was also adopted in other works such as \cite{dai-etal-2019-transformer}, as well as the Conformer paper \cite{gulati20_interspeech}, which is one of the state-of-the-art models for \gls{asr}. 
Previous works comparing absolute \glspl{pe} to relative ones for \gls{asr} come to the conclusion that especially for long speech inputs, absolute positional information is not as helpful as the relative counterpart \cite{9362093, Lu2019ATW}.

When performing computation with low precision, the absolute value distribution of each layer of the model plays an important role. 
Looking into the \gls{pe}, we found the output range to be different for encoding layers compared to regular layers.
This is especially relevant for specialized memristor hardware design, where the quantization range is predetermined by the hardware itself \cite{8528858}.
When using such consistent quantization settings, unwanted value clipping might occur.

\subsection{Contributions}

In this work, we investigate the behavior of a CTC-based Conformer with relative \gls{pe}, when being programmed and executed on simulated memristor hardware using SynaptogenML \cite{Synaptogen, rossenbach25_interspeech}. 
The core contributions are:
\begin{itemize}
	\item We confirm that relative \gls{pe} improves the model performance. This improvement is stronger for low-precision computation, e.g. using 4-bit weights.
	However, we observe degradation when adding \gls{pe} for memristor-based execution.
	\item We analyze the qualitative differences in the two networks, showing that the output value range of the \gls{pe} is not well captured by the default memristor configuration.
	\item We restore the performance gain of \gls{pe} using adapted model and device configurations. We improve by $\sim 15\%$ relative over the case without \gls{pe}, reducing the relative degradation during memristor device execution by $\sim 50$\%.
\end{itemize}
Furthermore, we discuss the implications of the different solutions we investigate in this work in the context of real world deployment scenarios.
The information can help researchers from both the hardware and the modeling community, when aiming to improve energy-efficient computing through application of memristive hardware.
All experiments can be run on a single GPU with 24 GB VRAM.
SynaptogenML, training code, recipes and other software \cite{DBLP:conf/emnlp/PeterBN18, DBLP:conf/icassp/DoetschZVKSN17, Wiesler-2014-RASRNNTheRWTHne} are publicly available.
\footnote{\url{https://github.com/rwth-i6/returnn-experiments/tree/master/2026-memristor-pe}}

\section{Background}

\subsection{Memristors}
Memristors are physical circuit elements that change their resistance through the application of a sufficiently large voltage or an electric field~\cite{Strukov2008}.
The resistance states within the device are non-volatile, meaning they are held until the next programming cycle.
The major benefit of memristors for machine learning is the capability to provide an analog interface for \gls{vmm} directly in memory~\cite{Li2018}. 
Through connecting multiple devices on crossbar arrays, each memristor can represent one or multiple bits of an entry in a weight matrix.
The input tensor of the operation can be converted into a voltage through \gls{dac} and applied to such an array. 
The resulting currents represent the \gls{vmm} output and are converted back into the digital domain using \gls{adc}.

Unfortunately, the circuits for \gls{dac} and \gls{adc} may require a large amount of energy and chip area compared to the actual computation network.
Adding them to nanometer-scale circuits still poses relevant challenges, and the precision range is usually limited to 8-bit in order to design realistic circuits \cite{8528858}.
This means that in addition to the uncertainty in the computation itself due to variation in the resistance states \cite{9091094}, the computation results are limited to a narrow digital precision.
Due to the necessity of physically storing weights and a slow programming process \cite{8306516}, only network layers with a static weight matrix, such as linear or convolutional layers, are currently feasible.

Chips integrating memristor technology are not available at a scale required for executing larger neural networks yet, so most experiments are run on small models for datasets such as MNIST~\cite{lecun-mnisthandwrittendigit-2010}, CIFAR~\cite{Krizhevsky09learningmultiple} or Google Speech Commands~\cite{speechcommandsv2}.
In order to overcome these limitations, we use a hardware simulation \cite{rossenbach25_interspeech} integrated into PyTorch based on the physically accurate device model Synaptogen \cite{Synaptogen}.
In its current state, the software allows to map a full Conformer network onto the simulated memristor devices.
For training, mapping and inference we refer to the public setup and recipes described in our previous work \cite{rossenbach25_interspeech}.

\subsection{Relative Positional Encodings}
\label{sec:relpos}
\glsunset{pe}
Relative \glspl{pe} have made widespread use for Transformer- and Conformer-based models.
While absolute encodings model the absolute position in the sequence, relative encodings model the relative distances between different positions.
In this work we stay conceptually close to the relative \gls{pe} as described in \cite{dai-etal-2019-transformer} and base our implementation on ESPNet\footnote{\url{https://github.com/espnet/espnet/blob/b74395b4fca4377e46cda7794e7a2e33bf644590/espnet2/asr_transducer/encoder/modules/attention.py}}.
This results in different possibilities in configuration, which we are also going to experiment with in this work.

The initial positional encoding matrix can either be based on a combination of sine and cosine functions \cite{vaswani2017attention, dai-etal-2019-transformer}, or can be learned during training \cite{shaw2018self}.
After this, the computed values are fed through a learned linear layer, which does not include a learnable bias.
Instead, two different biases are added to the query, one for the multiplication with the key and one for the multiplication with the encodings. 
During memristor execution the linear layer, which transforms the positional encodings as input to attention, is mapped onto the simulated devices.
\section{Experimental Setup}
\label{sec:exp_set}
We make use of a Conformer \cite{gulati20_interspeech} \gls{asr} system with \gls{ctc}-based \cite{10.1145/1143844.1143891} outputs. 
We conduct the majority of our studies on LibriSpeech \cite{librispeech} and verify our findings on Loquacious as a second task using the 250 hour subset for training.
For LibriSpeech, we use ARPA-phonemes from the provided lexicon as target labels.
For Loquacious, we use byte-pair-encoding \cite{DBLP:conf/acl/SennrichHB16a} with 128 merges.
As the memristor simulation requires a substantial amount of computation, we limit our evaluations to the \textit{dev-other} subset for LibriSpeech and \textit{dev} for Loquacious.

In recognition, we use a 4-gram count-based language model (LM) with KenLM \cite{heafield-2011-kenlm}, which is either the official LibriSpeech LM or the Loquacious LM from \cite{rossenbach2025supplementaryresourcesanalysisautomatic}.
Our models are being trained for 100 epochs. We use a linearly increasing and decreasing learning rate schedule peaking at $5 \cdot 10^{-4}$, with RAdam \cite{DBLP:conf/iclr/LiuJHCLG020} as optimizer and a decoupled weight decay of $1 \cdot 10^{-2}$.

We train our model using log-mel features with 10ms frame shift and downsample the frames by a factor of 4, through a convolutional front-end.
For regularization we use SpecAugment \cite{Park2019SpecAugmentAS}.
The Conformer consists of 12 layers, with model dimension 512 and feed-forward dimension 2048, with relative \gls{pe} added to the self-attention computation.
As described in \Cref{sec:relpos}, we use sinusoidal inputs transformed with a linear layer.
The total number of parameters of the models is $\sim 77$M.

We base our quantization-specific training settings and memristor mapping on the setup of our previous work in \cite{rossenbach25_interspeech}, where we give a detailed description of the configuration.
We place observers before and after the matrix-based operations and enforce symmetric quantization.
The observers are used to calculate the mapping to the memristive devices.
Activations will be quantized with 8-bit precision, while weights are quantized either to 8 or 4-bit precision.
In practice lower weight precision is desirable, as it reduces the overall number of required devices. 

Following the newest version of the SynaptogenML\footnote{\url{https://github.com/rwth-i6/SynaptogenML}} framework, we map all matrix operations with static weights of the Conformer model onto the simulated devices.
For more details on mapping, programming and execution of the simulated devices we refer to \cite{rossenbach25_interspeech}.
Each matrix operation is subdivided into computations of $128 \times 128$.
Within the sub-matrices, each bit level of the weight will be represented by a single crossbar array.
Each of the crossbars has its own \gls{adc}, before bit shifting and accumulation occurs.

When configuring the \gls{adc} in the simulation, the number of bits for precision and range of the conversion needs to be specified.
While the precision defines the decimal resolution, the range scales the absolute output value range, meaning the output is captured as a fixed-point value.
For \gls{adc} conversion our baseline uses the settings considered reasonable in \cite{rossenbach25_interspeech}, namely 4 bits for precision and 4 bits for range.
To keep the number of simulation runs feasible, we run 5 recognitions per model, each with differently programmed devices.
We report the average and the standard deviation of the programming runs.

\section{Experiments}

\subsection{Baseline}
\begin{table}
	\centering
	\caption{Results comparing baseline recognition and memristor execution, \gls{adc} is configured to 4-bit precision and 4-bit range. Activations are quantized to 8-bit.}
	\label{table:baseline}
	\vspace{-2ex}
	\resizebox{3.2in}{!}{
		\begin{tabular}{|c|c|c|c|c|c|}
			\hline
			\multirow{3}{*}{\makecell{Weight \\ Prec.}}&\multirow{3}{*}{\makecell{\gls{pe}}}& \multicolumn{4}{c|}{WER [\%]} \\
			\cline{3-6}
			& & \multicolumn{2}{c|}{Librispeech dev-other} &  \multicolumn{2}{c|}{Loquacious dev} \\
			\cline{3-6}
			& & Base. & Memristor & Base. & Memristor\\
			\hline
			\hline
			\multirow{2}{*}{8} & No&5.7 & \textbf{7.2} $\pm$ 0.07 & 13.1 & \textbf{14.5} $\pm$ 0.05 \\		
			\cline{2-6}
 			& \multirow{1}{*}{Yes}& \textbf{5.4} & 7.6 $\pm$ 0.08 & \textbf{12.7} & 16.4 $\pm$ 0.13 \\
			\hline
			\hline
			\multirow{2}{*}{4} &No&  6.5 & \textbf{7.5} $\pm$ 0.07 & 13.8 & \textbf{15.4} $\pm$ 0.10 \\
			\cline{2-6}
			& Yes& \textbf{5.6} & \textbf{7.5} $\pm$ 0.10 & \textbf{12.8} & 15.9 $\pm$ 0.20 \\
			\hline
	\end{tabular}
}
\vspace{-3.5ex}
\end{table}

For our first comparison, we run the pipeline for models with and without \glspl{pe}. The results can be seen in \Cref{table:baseline}. 
For the baseline recognition of the models without \glspl{pe} we can see that the model suffers from performance degradation when quantizing the weights down to 4 bits. 
Mapping the model onto the simulated devices yields another performance degradation from 10 to 25 \% relative WER, which is in a similar range as the one reported in \cite{rossenbach25_interspeech}.
The slight differences in degradation can be explained by the fact that the mapping now includes convolution and the model is about twice the size.
For the models with \gls{pe}, we can see that not only the 8-bit baseline outperforms the one without \gls{pe}, but also the 4-bit models are much more stable to the quantization. 
This means that the positional information helps the models improve especially in the low-precision scenario.
Yet, when executing the trained models on the simulated devices, we can see that the performance degradation is higher (25 to 40\% relative). 
Also the absolute performance does not beat the models without \gls{pe}. 
Given the previous results in \cite{rossenbach25_interspeech}, we expected that a performance increase by adding \gls{pe} to the baseline model would also improve the memristor performance.

\begin{table}[h]
	\centering
	\caption{Comparison of different \gls{adc} configurations for the full model on Librispeech dev-other. Activations are quantized to 8-bit. Oracle refers to keeping the encoding computation in the digital domain, i.e. not mapping it to the memristor devices.}
	\label{table:adc_variations}
	\vspace{-2ex}
	\begin{tabular}{|c|c|c|c|c|}
		\hline
		\multirow{3}{*}{\makecell{Positional \\ Encoding}}& \multicolumn{2}{c|}{\gls{adc}} & \multicolumn{2}{c|}{WER [\%]} \\
		\cline{2-5}
		& \multirow{2}{*}{\makecell{Prec.}} & \multirow{2}{*}{\makecell{Range}} &\multirow{2}{*}{\makecell{8 - bit \\ weights}} & \multirow{2}{*}{\makecell{4 - bit \\ weights}} \\
		&&&&\\
		\hline
		\hline
		\multirow{5}{*}{Yes} & \multirow{2}{*}{4} & \multirow{1}{*}{4} & \phantom{0}7.6 $\pm$ 0.08 & \phantom{0}7.5 $\pm$ 0.10  \\
		\cline{3-5}
		&  & 8 & \phantom{0}\textbf{6.6} $\pm$ 0.06 & \phantom{0}\textbf{6.8} $\pm$ 0.11\\
		\cline{2-5}
		& 8 & 8  & \phantom{0}7.0 $\pm$ 0.09 & \phantom{0}\textbf{6.9} $\pm$ 0.08 \\
		\cline{2-5}
		& 3 &5 & \phantom{0}7.3 $\pm$ 0.07 & \phantom{0}7.8 $\pm$ 0.04 \\
		\cline{2-5}
		& 2&6 & 38.9 $\pm$ 1.11 & 57.8 $\pm$ 0.94 \\
		\hline
		\hline
		\makecell{Oracle}& 4 & 4 & \phantom{0}\textbf{6.6} $\pm$ 0.09 & \phantom{0}\textbf{6.8} $\pm$ 0.06\\
		\hline
		\hline
		
		\multirow{3}{*}{\makecell{No}} & \multirow{2}{*}{4}&4 & \phantom{0}7.2 $\pm$ 0.07 & \phantom{0}7.5 $\pm$ 0.07 \\
		\cline{3-5}
		& &8 & \phantom{0}7.2 $\pm$ 0.06 & \phantom{0}7.5 $\pm$ 0.11 \\
		\cline{2-5}
		& 8&8 & \phantom{0}8.3 $\pm$ 0.11 &\phantom{0}7.6 $\pm$ 0.08 \\
		\hline
	\end{tabular}
	\vspace{-4.5ex}
\end{table}

\subsection{ADC Variations}
\label{sec:adc}
\begin{table*}[t]
	
	\centering
	\caption{Final Results comparing the best approaches for degradation mitigation. Activations are quantized to 8-bit. The \gls{adc} of non-\gls{pe} parts of the model is configured to 4-bit precision and 4-bit range. Oracle refers to keeping the encoding computation in the digital domain, i.e. not mapping it to the memristor devices.}
	\label{table:final}
	\vspace{-2ex}
	\begin{tabular}{|c|c|c||c|c|c|c||c|c|c|c|}
		\hline
		\multirow{4}{*}{\makecell{Positional \\ Encoding}}& \multicolumn{2}{c||}{\multirow{2}{*}{\makecell{\gls{pe} ADC}}}& \multicolumn{8}{c|}{WER [\%]} \\
		\cline{4-11}
		& \multicolumn{2}{c||}{} & \multicolumn{4}{c||}{LibriSpeech dev-other} & \multicolumn{4}{c|}{Loquacious dev} \\
		\cline{2-11}
		& \multirow{2}{*}{Prec.} &\multirow{2}{*}{Range}&\multicolumn{2}{c|}{8 - bit weights} & \multicolumn{2}{c||}{4 - bit weights} & \multicolumn{2}{c|}{8 - bit weights} & \multicolumn{2}{c|}{4 - bit weights}\\
		\cline{4-11}
		& && Base. & Memristor & Base. & Memristor & Base. & Memristor & Base. & Memristor \\
		\hline
		\hline
		No& -  & - & 5.7 & 7.2 $\pm$ 0.07& 6.5 & 7.5 $\pm$ 0.07 &13.1&14.5 $\pm$ 0.05 &13.8&15.4 $\pm$ 0.10 \\
		\hline
		\hline
		\multirow{3}{*}{Yes} & 4 & 4 & \multirow{3}{*}{5.4} & 7.6 $\pm$ 0.08 & \multirow{3}{*}{5.6} & 7.5 $\pm$ 0.10 & \multirow{3}{*}{12.7} &  16.4 $\pm$ 0.13 & \multirow{3}{*}{12.8}&15.9 $\pm$ 0.20 \\
		\cline{2-3}
		\cline{5-5}
		\cline{7-7}
		\cline{9-9}
		\cline{11-11}
		&4&8& &6.6 $\pm$ 0.06&& 6.8 $\pm$ 0.06&&14.2 $\pm$ 0.04&&14.1 $\pm$ 0.07\\
		\cline{2-3}
		\cline{5-5}
		\cline{7-7}
		\cline{9-9}
		\cline{11-11}
		& 1 & 7&&6.5 $\pm$ 0.06 &&6.8 $\pm$ 0.08&&14.1 $\pm$ 0.06&&14.1 $\pm$ 0.02\\
		\cline{1-11}
		No Linear & - & - & 5.6 & 6.9 $\pm$ 0.06 & 5.8 & 6.8 $\pm$ 0.07&12.9& 14.6 $\pm$ 0.07 &13.0&14.3 $\pm$ 0.06\\
		\hline	
		\hline		
		Oracle& - & - &\multirow{1}{*}{5.4}&6.6 $\pm$ 0.09 &\multirow{1}{*}{5.6} &6.8 $\pm$ 0.06&\multirow{1}{*}{12.7}&14.2 $\pm$ 0.08&\multirow{1}{*}{12.8}&14.1 $\pm$ 0.05\\
		\hline
	\end{tabular}
\end{table*}

As the only difference between the two types of models lies in the addition of \glspl{pe}, we will now investigate them in more detail.
As mentioned in \Cref{sec:exp_set}, the output value range of the default \gls{adc} conversion setting is bound by 4 bits.
This is below the 8-bit value range that is theoretically reachable by the 128x128 \gls{vmm} with binary weights.
To verify that this is the issue, we track the \gls{adc} operation of the linear layer of the \gls{pe} for each bit and device during a forward through the network.
On average around 40 \% of the time the result of the memristor computation is being clipped, with almost 100\% for some crossbars in extreme cases. 
This causes a major shift in the output distribution of the \gls{pe}.
To get oracle results, we perform recognition runs for the models, where we do not map the linear layer of the \gls{pe} onto the devices, but rather compute them digitally as usual.
In \Cref{table:adc_variations}, the oracle results show clear improvements over the baseline, almost halving the overall degradation caused by the memristor mapping.
The \gls{adc} and \gls{dac} settings were not investigated by the authors in \cite{rossenbach25_interspeech}.
This means that it might also be an issue in other portions of the model.
Because of this, we investigate multiple different configurations with respect to precision and range of the \gls{adc}.

For both 8 and 4-bit weight precision, increasing the range to 8-bit shows a clear improvement over the baseline setting.
Interestingly, increasing the precision of the conversion improves the performance, but does not keep up with only increasing the range. 
A possible explanation for this is that a higher precision will enhance the memristor noise, while lower precision \glspl{adc} quantize some of it away. 
As this kind of \gls{adc} analysis is novel, we also run these experiments on the models without \gls{pe}, visible in the last rows of \Cref{table:adc_variations}.
Not only does the increased range not help in this case, but the increased precision even degrades the model for 8-bit precision. 
From this we conclude that the other parts of the network do not suffer from the same range issues as the \gls{pe} does. 

\begin{figure}
	\caption{Comparison of the different \gls{adc} settings for the \gls{pe} of the 8-bit weight model. The \gls{adc} of other parts of the model is set to 4-bit range and 4-bit precision.}
	\label{fig:posadc}	
	\vspace{-1.0em}
	\centering
	\includegraphics[width=\linewidth, clip, trim={15pt 15pt 0 10pt}]{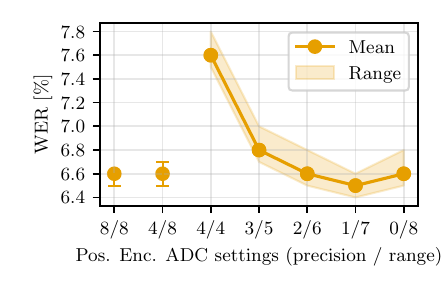}
	\vspace{-3.0em}
\end{figure}

In the next set of experiments we only change the settings of layers influencing the \gls{pe}, but keep the rest of the network settings fixed at the default 4-bit range and precision. 
The first two values in \Cref{fig:posadc} show that in this case increasing the output range improves the model's performance, reaching the same optimum as an increase for the whole model.
When also increasing the precision of the conversion, the model shows more robustness to the increased level of noise and does not degrade. 

While studies with arbitrarily configured \glspl{adc} based on simulation are possible, in a real-life deployment setting, increasing the range or precision of the \gls{adc} might not be feasible.
The increased bit levels will drastically increase the overall energy and area consumption of the \gls{adc} component. 
The result of this could be that the energy savings of the memristors might be lost in conversion. 
Because of this, we keep the overall bit budget fixed at 8 bits, and only shift bits from the precision towards the range. 
When applying the fixed budget to the full model in \Cref{table:adc_variations}, we see minor improvements over the baseline for a precision/range split of 3/5.
Below that, a strong degradation sets in.
This means that the precision bits are required at least for some parts of the model.
In contrast when only changing the conversion of the \gls{pe}, we can see in \Cref{fig:posadc} a performance gain up to a range of 7 or 8 bits.
Note that 0 precision bits imply that no fractional values are allowed, but a value range up to 128 integers is possible.
While the best result is using 7 bits of range, we again assume that the difference between 6 to 8 bits in range reflects the uncertainty of the devices rather than an actual performance difference.

\subsection{Pos Enc Variations}
\label{sec:posenc}

In physical circuits, dynamic changes to the \gls{adc} might be difficult to realize under desired space and power constraints.
This is why we investigate whether we can improve the model itself in preparation for memristor deployment.
We implement multiple changes to the \gls{pe}, which can be seen in \Cref{table:pos_enc_types}. 
One reason for the deviation of layer statistics in the \gls{pe} transformation layer might be the artificial structure of sinusoidal encodings used as input.
Because of this, we replace them by learned \gls{pe} as initially proposed in \cite{shaw2018self}.

The baseline performance, compared to the default configuration, is worse by 0.2\% WER absolute.
Yet, when mapping the model onto the devices, one can see that the degradation is slightly lower, especially for 4-bit weight precision. 
While this indicates that a part of the issue is caused by the sinusoidal encodings, it does not account for all of the degradation.

Secondly, we experiment with retraining the model without the linear transformation.
This is different from the oracle case, where we just do not map the layer onto the devices.
From the baseline result one can see that the performance of the model degrades by 0.2\% WER absolute. 
Yet, as there is no linear to map to the devices, the overall performance during memristor execution surpasses the default configuration, only staying behind larger \gls{adc} ranges and the oracle performance. 
This means that in a setting where the \gls{adc} cannot be modified, this approach is a reasonable alternative.
As the \gls{pe} component in this case has no learnable parameters, we also try swapping the sinusoidal encodings for trainable ones.
While the baseline performance stays about the same as for the sinusoidal encodings without a linear, the memristor execution is slightly worse. 
We conclude that the learnable encodings themselves are not able to compensate for the missing linear layer.

\subsection{Final Results}
\begin{table}[]
	
	\centering
	\caption{Comparison of modifications to the relative \gls{pe}. \gls{adc} is configured to 4-bit precision and 4-bit range. Activations are quantized to 8-bit. Oracle refers to keeping the encoding computation in the digital domain, i.e. not mapping it to the memristor devices.}
	\label{table:pos_enc_types}
	\vspace{-2ex}
	\begin{tabular}{|c|c|c|c|c|}
		\hline
		\multirow{3}{*}{\makecell{Positional \\ Encoding}}& \multicolumn{4}{c|}{WER [\%]} \\
		\cline{2-5}
		& \multicolumn{2}{c|}{8 - bit weights} & \multicolumn{2}{c|}{4 - bit weights} \\
		\cline{2-5}
		& Base. & Memristor & Base. & Memristor \\
		\hline
		\hline
		No&  5.7 & 7.2 $\pm$ 0.07 & 6.5 & 7.5 $\pm$ 0.07 \\		
		\hline
		Baseline & \textbf{5.4} & 7.6 $\pm$ 0.08 & \textbf{5.6} & 7.5 $\pm$ 0.10 \\
		\hline
		Oracle& \textbf{5.4} & \textbf{6.6} $\pm$ 0.09 & \textbf{5.6} & \textbf{6.8} $\pm$ 0.06\\
		\hline
		\hline
		Learnable & 5.7 & 7.4 $\pm$ 0.14 &  5.8 & 7.0 $\pm$ 0.05\\
		\hline
		No Linear & 5.6 & \textbf{6.9} $\pm$ 0.06 & 5.8 & \textbf{6.8} $\pm$ 0.07 \\	
		\hline
		\makecell{~~~+ Learn.} &5.5 & 7.0 $\pm$ 0.10 & 5.8 & 7.0 $\pm$ 0.06\\
		\hline
	\end{tabular}
	\vspace{-4ex}
\end{table}
The final results can be found in \Cref{table:final}. 
For this we use the best performing configurations from the ablation studies in \Cref{sec:adc} and \Cref{sec:posenc} and run them on Loquacious to verify our findings.
While the overall error range of the corpus is higher, the models show a similar trend. 
Increasing the precision of the \gls{adc} helps both for the complete model, as well as only for the mapping of the \gls{pe}. 
Shifting the precision bits into the range of the computation of the mapped encodings brings a similar improvement, while keeping the estimated energy consumption constant. 
One minor difference is the fact that the 4-bit model seems to handle the memristor execution slightly better than for LibriSpeech, as the performance is marginally better than the 8-bit case. 
A possible explanation for this might be the fact that the overall error rates are higher and the model already has to deal with increased uncertainty.
As for LibriSpeech, dropping the linear mapping degrades the baseline model slightly, but produces a better model in the default memristor configuration.
Interestingly, the memristor performance for the 8-bit weights does not surpass the model without \gls{pe}, while for the lower bit precision the importance of the \gls{pe} shows once again.
Overall, we can conclude that our changes improve the memristor execution mostly independent of the dataset.
\section{Conclusion}
In this work, we investigated the issues that can arise when mapping a state-of-the-art Conformer model with relative \gls{pe} onto simulated memristor devices.
We identified the encodings to require specific care, as they do not integrate into the memristor execution as smoothly as the rest of the network. 
In order to mitigate the degradation caused by the increased noise and reduced output range of the memristor devices, we investigated a number of methods.
Increasing the range bits of the \gls{adc} reduces the degradation by $50$\% relative, while changing the model architecture reduces it by $30$\%.
Which method is applicable strongly depends on the use case and scenario.
In a situation where the chip used for execution is still configurable in its \gls{adc} settings, or statically provides different configurations, a higher number of range bits restores the performance gain of the \gls{pe}.
Another possible scenario is where the chip configuration does not provide multiple levels and is static. 
There, removing the linear mapping from the model before training degrades the baseline performance slightly, but results in a better performance during memristor execution.

\section{Acknowledgments}
This work was partially supported by NeuroSys, which as part of the initiative “Clusters4Future” is funded by the Federal Ministry of Research, Technology and Space BMFTR (funding IDs 03ZU2106DA and 03ZU2106DD), and by the project RESCALE within the program \textit{AI Lighthouse Projects for the Environment, Climate, Nature and Resources} funded by the Federal Ministry for the Environment, Nature Conservation, Nuclear Safety and Consumer Protection (BMUV), funding ID: 67KI32006A.
The authors gratefully acknowledge the computing time provided to them at the NHR Center NHR4CES at RWTH Aachen University (project number p0023999).
This is funded by the Federal Ministry of Education and Research, and the state governments participating on the basis of the resolutions of the GWK for national high performance computing at universities (\url{www.nhr-verein.de/unsere-partner}).
We would like to thank Leon Brackmann and Stephan Menzel for their helpful input and discussions on the design and verification of the simulation framework.

\section{Generative AI Use Disclosure}
Generative AI was used in this work for proofreading and grammar correction, but was not used to generate content.

\bibliographystyle{IEEEtran}
\bibliography{mybib}

\end{document}